\documentclass[10pt,twocolumn,letterpaper]{article}

\usepackage[pagenumbers]{cvpr} 

\usepackage{graphicx}
\usepackage{amsmath}
\usepackage{amssymb}
\usepackage{booktabs}

\usepackage{latexsym}
\usepackage{times}
\usepackage{colour-blind}
\usepackage{url}
\usepackage{color}
\usepackage{graphicx}
\definecolor{gray}{RGB}{100,100,100}

\newcommand{\tightset}[1]{\{#1\}}

\newcommand{\R}[0]{\mathbb{R}}
\newcommand{\para}[1]{\noindent\textbf{#1}}

\usepackage[hang,flushmargin]{footmisc} 
\usepackage[pagebackref,breaklinks,colorlinks]{hyperref}

\usepackage[capitalize]{cleveref}
\crefname{section}{Sec.}{Secs.}
\Crefname{section}{Section}{Sections}
\Crefname{table}{Table}{Tables}
\crefname{table}{Tab.}{Tabs.}

\def\cvprPaperID{3416}
\def\confName{CVPR}
\def\confYear{2022}

\begin{document}

\title{Revisiting the ``Video'' in Video-Language Understanding}

\author{
Shyamal Buch$^{1}$, Crist\'{o}bal Eyzaguirre$^{1}$, Adrien Gaidon$^{2}$, Jiajun Wu$^{1}$, Li Fei-Fei$^{1}$, Juan Carlos Niebles$^{1}$\\
$^{1}$Stanford University, $^{2}$Toyota Research Institute\\
{\tt\small \{shyamal, ceyzagui, jiajunwu, feifeili, jniebles\}@cs.stanford.edu, adrien.gaidon@tri.global}
}
\maketitle

\begin{abstract}
What makes a video task uniquely suited for \textit{videos}, beyond what can be understood from a single image?
Building on recent progress in self-supervised image-language models, we revisit this question in the context of
video and language tasks.
We propose the atemporal probe (ATP), a new model for video-language analysis which provides a stronger bound on the baseline accuracy of multimodal models constrained by image-level understanding.
By applying this model to standard discriminative video and language tasks, such as video question answering and text-to-video retrieval,
we characterize the limitations and potential of current video-language benchmarks. We find that understanding of event temporality is often not necessary to achieve strong or state-of-the-art performance, even compared with recent large-scale video-language models and in contexts intended to benchmark deeper video-level understanding.
We also demonstrate how ATP can improve both video-language dataset and model design. We describe a technique for leveraging ATP to better disentangle dataset subsets with a higher concentration of temporally challenging data, improving benchmarking efficacy for causal and temporal understanding.
Further, we show that effectively integrating ATP into full video-level temporal models can improve efficiency and state-of-the-art accuracy.%
\footnote{Project website: \href{https://stanfordvl.github.io/atp-revisit-video-lang/}{https://stanfordvl.github.io/atp-revisit-video-lang/}}
\end{abstract}



\section{Introduction}
\label{sec:intro}
\begin{figure}[ht]
\begin{center}
\centerline{\includegraphics[width=\columnwidth]{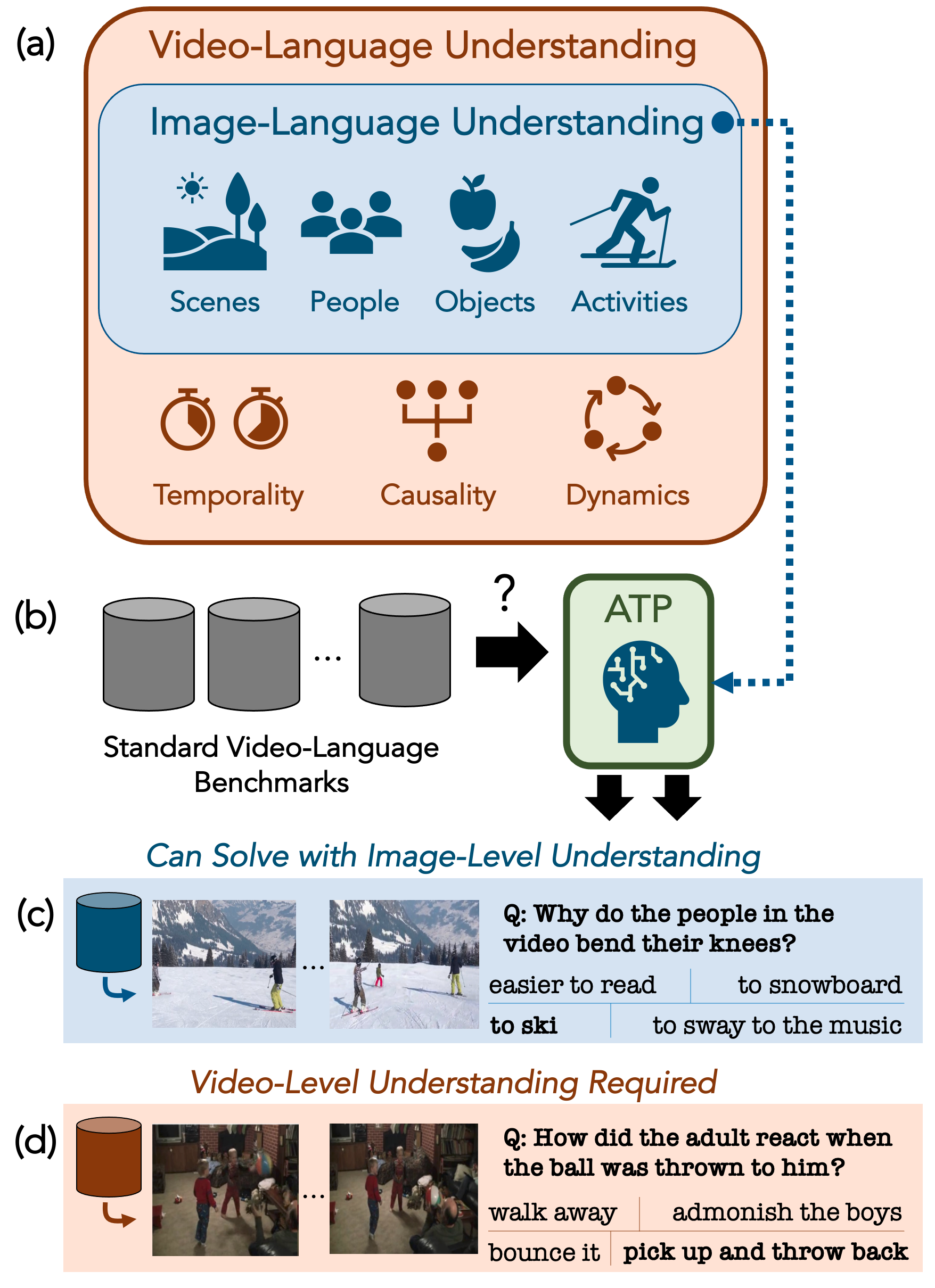}}
\vskip -0.15in
\caption{
	\textbf{(a)} The promise of \textit{videos} lies in the potential to go beyond \textit{image-level} understanding (scenes, people, etc.) to capture event temporality, causality, and dynamics. 
    \textbf{(b)} In this work, we propose an atemporal probe (ATP) model to \textit{revisit the video} in standard benchmarks \cite{xiao2021next,li2021value,xu2016msr} for 
    video question answering and text-to-video retrieval, offering a stronger image-centric baseline 
    and analytical tool. For example, ATP finds non-trivial subsets of ``causal'' questions that can be answered with \textbf{(c)} only image-level understanding, rather than \textbf{(d)} full video-level understanding.
}
\label{fig:pull-fig}
\end{center}
\vskip -0.35in
\end{figure}

Videos offer the promise of understanding not only what can be discerned from a single image (e.g. scenes, people, and objects), but also multi-frame event temporality, causality, and dynamics (Figure~\ref{fig:pull-fig}(a)). 
Correspondingly, there lies a central question at the heart of video research:
\textit{What makes a video task uniquely suited for videos, beyond what can be understood from a single image?}

As a field, video analysis has considered this question deeply in the context of action classification in videos \cite{schindler2008action,wang2016temporal,carreira2017quo,huang2018what}. The emergence of strong convolutional models for image classification \cite{he2016deep} enabled researchers to better characterize the limits of single-frame understanding for recognizing actions \cite{wang2016temporal,huang2018what}. A key finding from this analysis was that, in many standard video datasets \cite{soomro2012ucf101,kuehne2011hmdb} at the time, temporal understanding was simply not required to perform well on these benchmarks. For example, recognizing static scene context like the presence of a pool was sufficient to recognize the ``diving'' activity from a single frame \cite{liu2021no,wang2016temporal}. The impact of such analysis was tremendous: later datasets were designed to capture a richer distribution of temporal understanding \cite{goyal2017something,diba2020large,smaira2020short}
with better disentanglement of such cues \cite{lu2021metadata}, and model designs evolved further to better capture the now necessary dynamics to address these improved tasks \cite{lin2019tsm,fanbuch2020rubiksnet,feichtenhofer2019slowfast,feichtenhofer2020x3d,lfb2019}.

Meanwhile, the recent advent of self-supervised image-language models \cite{radford2018improving,jia2021scaling} with competitive performance to standard image-classification models \cite{he2016deep,dosovitskiy2020image} means that we have a unique opportunity to reconsider this fundamental question in the context of standard discriminative \textit{video-language} tasks, such as video question answering \cite{xiao2021next,li2021value,xu2016msr} and video-language retrieval \cite{xu2016msr,hendricks2018localizing,krishna2017dense}.
In particular, we can now build \textit{beyond} prior (video-only) analysis work, largely constrained to recognition settings of limited atomic actions in relatively short clips, towards more complex (temporal, causal) event understanding in longer-horizon, multimodal settings where the expressivity of natural language can potentially describe a richer event space.

The primary motivation of our work is to analyze these existing video-language benchmarks by \textit{revisiting the video}, and derive insights that can help guide the further development of the field. Our driving question is, to what extent can image-level understanding obtained from a single frame (well-chosen, without temporal context) address the current landscape of video-language tasks? To accomplish this, we make the following key contributions:

First, we introduce the atemporal probe (ATP) model to provide a stronger bound on the capabilities of image-level understanding in video-language settings than traditional random frame and mean pooling baselines \cite{wang2016temporal}. Here, we leverage a \textit{frozen} self-supervised image-language model (e.g. CLIP \cite{radford2021learning}) to extract a set of image and language representations: our ATP model must then learn to select a \textit{single} frozen representation corresponding to a single frame, and forward that to the downstream video-language task. Critically, our framework is constrained to \textit{not} be capable of reasoning temporally, and its output is ultimately bottlenecked by what a frozen image-language model can discern from an individual, decontextualized video frame.

Second, we apply ATP to analyze a wide range of video-language datasets, focusing primarily on video question answering with extensions to text-to-video retrieval (per Figure~\ref{fig:pull-fig}(b)). To our surprise, we find that many standard and recent benchmarks can be potentially well-addressed with single-frame image understanding. In particular, while this was not our primary aim, we find that our \textit{learned} ATP model is able to outperform recent state-of-the-art video-language models on standard vision-language benchmarks \cite{xiao2021next,li2021value,xu2016msr}, despite its substantial bottleneck constraints on model capacity, capability, and inputs. We find that even recent benchmarks that explicitly design for temporal and causal understanding (e.g., \cite{xiao2021next}), can have a non-trivial subset of questions answerable by simple single-frame event recognition. As shown in Figure~\ref{fig:pull-fig}(c), 
while the question asking ``why'' an event occurred suggests causal understanding may be needed, our ATP model shows that in practice simple scene and object recognition can ascertain the correct answer from a single chosen frame.

Finally, we examine how ATP and the insights it provides can help with improving both dataset and video-level temporal modeling designs. As a case study, we closely examine the NExT-QA benchmark \cite{xiao2021next}. We find that ATP is able to better identify collections of ``causal'' and ``temporal'' questions that \textit{cannot} be well-addressed with single-frame understanding. In Figure~\ref{fig:pull-fig}(d), ATP struggles to answer this question since it necessitates multi-event reasoning across time. By improving the disentanglement of video- and image-level understanding in the benchmark data, we can better understand the progress of state-of-the-art video techniques leveraging motion features and event reasoning architectures over image-centric models, a result that is not as apparent in the original setting. We further validate our analysis by training a temporal video-level model on top of our ATP selectors, achieving a new state-of-the-art for this benchmark with improved efficiency. Taken together, our analysis suggest key avenues by which our ATP technique can guide continued development of video-language datasets and models in future work.

\section{Background and Related Work}
\label{sec:related_work}

Our work is related to many different areas of vision and vision-language research, including video-specific and image-specific settings. In this section, we discuss the key relevant areas of prior work that motivate our contributions.

\para{Video-language understanding (tasks).} Understanding events in their multimodal vision-language context is a long-standing challenge for the computer vision community. Standard video-language tasks include both discriminative tasks, such as
video question answering \cite{xiao2021next,li2021value,xu2016msr,lei2018tvqa,lei2020vlep,jang2017tgif,yu2019activityqa,yi2019clevrer,girdhar2019cater},
text-to-video/moment retrieval \cite{xu2016msr,hendricks2018localizing,krishna2017dense,rohrbach2015dataset,zhou2017towards},
and generative tasks, such as video captioning \cite{krishna2017dense,chen2015microsoft} and open-ended VQA \cite{xiao2021next,xu2016msr}. In context, we choose a representative subset of these video-language benchmarks well-suited to studying event temporality and causality. In particular, we choose to focus on \textit{discriminative} tasks, since automatic metrics (without human-in-the-loop) for generative tasks with causal descriptions remains an open research challenge \cite{pillutla2021information}. Furthermore, many video-language tasks involve heavy reasoning over auxiliary text inputs, such as scripts \cite{yu2019activityqa,engin2021hidden}.
These exciting directions are complementary to our goal: we focus instead on revisiting event temporality in the real-world videos themselves.

\para{Video-language understanding (approaches).} Standard approaches for addressing these tasks \cite{li2020hero,jiang2020reasoning,krishna2017dense,xu2021videoclip,sun2019videobert,miech2020end} often operate on a combination of image-derived appearance \cite{he2016deep,dosovitskiy2020image} and video-derived motion features \cite{carreira2017quo,simonyan2014two,qiu2017learning,miech2019howto100m} as input to an architecture \cite{vaswani2017attention,zhou2018temporal} that combines information across the temporal dimension for the final task. While these models are traditionally quite heavy, employing dense features extracted from many frames, recent work \cite{lei2021less} has suggested that enabling end-to-end training through sparsity can improve accuracy. Our proposed approach aims to complement these prior lines of work by taking a different approach: instead of focusing explicitly on improving state-of-the-art accuracies, we impose strong learnability and representation constraints to better \textit{analyze} the degree to which full video-level understanding is truly necessitated by current benchmarks, to help guide future model and dataset designs for capturing deeper event understanding.

\para{Temporality in videos (action recognition).} Action and event recognition are fundamental tasks for video understanding, and the subject of recurring deep analysis regarding the role of temporality in action classification \cite{schindler2008action,wang2016temporal,carreira2017quo,huang2018what,zhou2018temporal}, with profound downstream impacts on dataset \cite{goyal2017something,diba2020large,smaira2020short} and subsequent model designs \cite{lin2019tsm,fanbuch2020rubiksnet,feichtenhofer2019slowfast,feichtenhofer2020x3d,lfb2019}. We draw inspiration from this foundational prior work, while also aiming to broaden analysis beyond characterizing limited sets of atomic actions towards longer-horizon temporal and causal event understanding, which multimodal video-language contexts have the potential to better capture \cite{xiao2021next}.

\para{Image-language understanding.} The advent of new self-supervised vision-language models trained at scale \cite{radford2021learning,jia2021scaling}, where models learn a joint embedding space for vision \cite{dosovitskiy2020image,he2016deep} and language \cite{liu2019roberta,devlin2018bert} \textit{without} explicit low-level labels, has proven disruptive for image and image-language understanding tasks \cite{shen2021much,radford2021learning,antol2015vqa}.
We leverage these models, both vision and language components, as foundations for our analytical technique to better characterize the extent to which image-language understanding can address current video-language tasks. Our work is \textit{complementary} to prior image-language analytical work \cite{goyal2017making} which revealed unintended language bias: we aim to characterize the extent of unintended video-specific biases in this multimodal setting.

\para{Efficient image-centric video modeling.} Finally, we note that aspects of our technical approach draw inspiration from efficient image-centric video modeling literature, which aim to improve efficiency and for tasks like action recognition \cite{wu2019adaframe} and localization \cite{yeung2016end} by learning how to selectively process a sparse number of frames from the input video.

\begin{figure}[t]
\begin{center}
\centerline{\includegraphics[width=\linewidth]{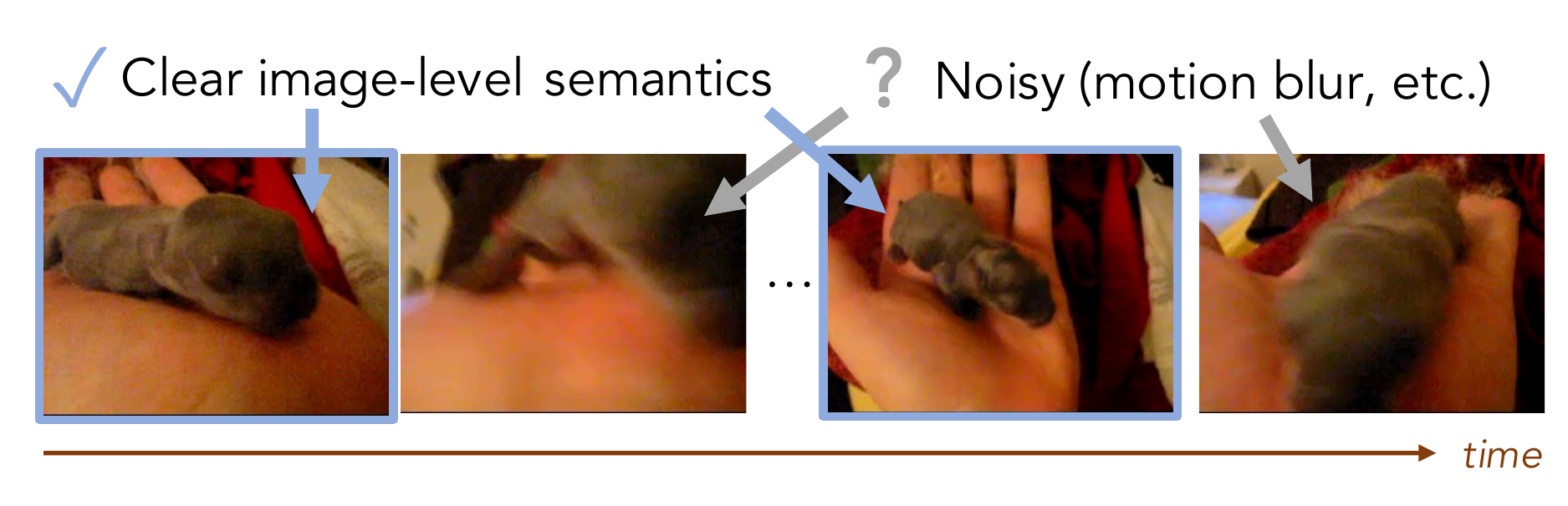}}
\vskip -0.1in
\caption{\textbf{Motivating a stronger image-centric baseline}. Videos are noisy, correlated collections of frames \cite{liu2021no}: while some frames have clear image-level semantics (\textit{above:} a small puppy dog in a human hand), a significant fraction of frames can contain camera motion blur, difficult perspectives, and uninformative frames. Standard atemporal techniques, such as evaluating image-level models on a random frame or mean pooling, may be susceptible to such noise, and thus not necessarily represent a true bound on image-level semantics understanding in video-language contexts. This motivates our atemporal probe (ATP) model (Sec.~\ref{sec:technical_approach}).
}
\label{fig:model-fig-motivation}
\end{center}
\vskip -0.2in
\end{figure}

\begin{figure*}[t]
\begin{center}
\centerline{\includegraphics[width=0.9\linewidth]{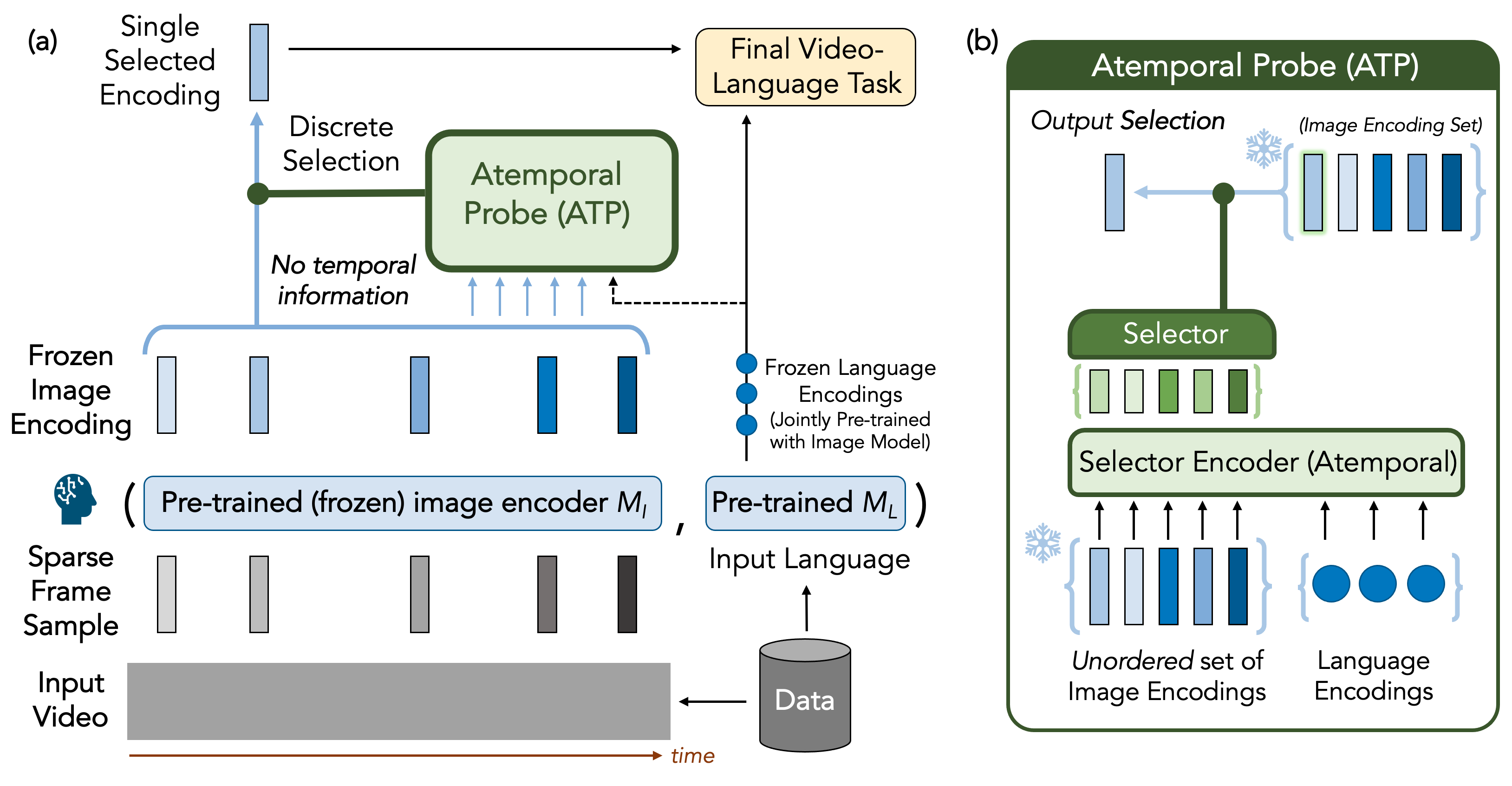}}
\vskip -0.2in
\caption{\textbf{Atemporal Probe (ATP).} We propose ATP: a new, stronger baseline for characterizing the degree to which video-language tasks can be addressed exclusively with vision-language understanding derived from image-only settings (i.e. jointly learned pre-trained encoders for image $M_I$ and language $M_L$). \textbf{(a)} In the broader context of a video-language task, such as video question answering, our ATP model must learn to select a \textit{single} (frozen, image-derived) embedding that can provide as strong a signal as possible for the final task. \textbf{(b)} Zooming in, we emphasize that our ATP model does not use any temporal information as part of this selection and is permutation-invariant, operating on an unordered (shuffled) set of frame-level embeddings (without temporal positional encodings) with self-attention operations. Furthermore, the learnable atemporal selector encoder remains low capacity. Please see Section~\ref{sec:ta:atp-model} for additional details.
}
\label{fig:model-fig-atp}
\end{center}
\vskip -0.3in
\end{figure*}

\section{Technical Approach}
\label{sec:technical_approach}

\noindent In this section, we describe our technical approach for our atemporal probe (ATP), a new modeling tool for characterizing the boundary of image-constrained understanding in the context of standard discriminative video-language tasks.

\subsection{Preliminaries: Video-Language Tasks}
\label{sec:ta:prelim-vl-tasks}

We first briefly introduce the notation and discriminative video-language tasks we consider in this work, namely video question answering and text-to-video retrieval:

\para{Video question answering.} Our primary analysis setting is on video question answering: given a paired collection of videos $C_V$, and language questions and answers $C_L = \tightset{C_Q, C_A}$, the goal is for each (video, question) pairing $(V, Q)$ to provide the correct answer in $A$.

\para{Video-language retrieval.} We also examine video-language retrieval, to assess the generality of our approach. In text-to-video retrieval, the objective is complementary: given a paired collection of videos $C_V$ and language descriptions $C_L$, the goal is to use the language $L$ to retrieve the specific video $V$ that it originally corresponded with.

We note that in both settings, there exist video $V$ and language $L$ ($= (Q, A)$) inputs common to each task. While our work ultimately analyzes performance on these downstream tasks with respect to their inputs and metrics, our core goal for this work is to provide an improved analytical tool for characterizing specific instantiations of these tasks.

\subsection{Motivating a Stronger Image-Centric Baseline}
\label{sec:ta:motivation}

Traditionally, video models and benchmarks establish their efficacy over image-level understanding by reporting results with a model based on a single (center-most, randomly, etc.) chosen video frame \cite{wang2016temporal}.
Because videos can be considered noisy collections of frames, such baselines may not truly represent the bounds of what image-constrained understanding can achieve in video-language contexts (Figure~\ref{fig:model-fig-motivation}).
In particular, we seek to answer the question: if we can select a ``good frame'' from the video and only derive our understanding from that one frame, what video-language tasks are we capable of performing?

Intuitively, settings where only scene-level descriptions are being assessed should likely be addressable from a single frame, as should simple event recognition (per prior analysis in the domain of action recognition, Section~\ref{sec:related_work}). However, by the same intuition, questions/tasks that attempt to fully assess deeper event dynamics, causal, or temporal understanding should in principle be \textit{unanswerable} from a single frame alone, requiring reasoning over multiple events which are not necessarily co-located in time. A compelling baseline that effectively bounds image-level understanding can thus potentially help distinguish between these settings.

\subsection{Atemporal Probe (ATP) Model}
\label{sec:ta:atp-model}

\para{Overview.} With the motivating insight above, we propose an atemporal probe (ATP) model: a new, stronger analytical approach for characterizing the degree to which video-language tasks can be addressed exclusively with vision-language representations derived from image-only settings. The ATP model (Figure~\ref{fig:model-fig-atp}) is tasked with finding a single (frozen, image-derived) embedding from the video and forwarding this to the downstream video-language task. Our ATP model does \textit{not} use any temporal information to perform this selection and is permutation-invariant, processing unordered frame embeddings with self-attention operations (without any sequence positional information). Further, we ensure that the learnable portion of ATP remains low capacity, with only a few, small layers and number of heads.

\para{ATP (Context).} We illustrate an overview of our ATP model in the larger video-language task context in Figure~\ref{fig:model-fig-atp}(a). For each video  $V \in C_{V}$, we draw a random sparse (shuffled) subset of frames $F = \tightset{v_1,\dots,v_n} \in V$, where usually $n << |V|$, the length of the video. We also take as input to our task a pretrained, self-supervised image-language model $M = \tightset{M_{I}, M_{L}}$, which consists of two components $M_{I}$ and $M_{L}$ for the vision and language components, respectively. These are used to encode all video $V$ and language $L$ inputs to the original video-language task.

We proceed to encode each of the frames with the pre-trained vision encoder $M_I(F) = \tightset{x_1,\dots,x_n}$ to get vision embeddings $x_i$ corresponding to each frame $v_i$. Intuitively, because our encoder is completely frozen and never updated, $x_i$ is a representation of what an \textit{image-constrained} visual encoder can discern; no additional information of the broader video is encoded here. Furthermore, our model treats the set $\tightset{x_1,\dots,x_n}$ as an \textit{unordered} set, without any temporal positional information.

Now, ATP can be properly formulated as:
\begin{equation}
    ATP: \tightset{x_1, \dots, x_n} \mapsto x_i,
\end{equation}
where the goal is to select a single representation $x_i \in \tightset{x_1, \dots, x_n}$ to pass to the final video-language task. Depending on the original video-language task formulation, ATP can take additional language inputs $M_{L}(L)$ (e.g. the encoded question for video question answering; Sec.~\ref{sec:exp:bench-implem}).

\para{ATP (Selection).} In Figure~\ref{fig:model-fig-atp}(b), we illustrate a more detailed view of the ATP selection operation. Given the inputs provided by the frozen pre-trained image and language encoders, the ATP model must now perform \textit{embedding selection}, passing one of these input visual embeddings, unmodified, to the downstream video-language task. To accomplish this, ATP first encodes the (unordered, shuffled) input image encoding sequence $\tightset{x_1,\dots,x_n}$ with a learnable selector encoder $E_s$ as follows:
\begin{equation}
    E_s(\tightset{x_1,\dots,x_n}; M_L(L)) \mapsto \tightset{s_1, \dots, s_n},
\end{equation}
where $\tightset{s_1, \dots, s_n}$ correspond to the original $\tightset{x_1,\dots,x_n}$ and are only used for selection. We instantiate $E_s$ in our work as low-capacity transformer \cite{vaswani2017attention}, with 3 or fewer layers and heads: we choose a self-attention based encoder here because it is conducive towards permutation invariant model design \cite{lee2019set}. Because our original embedding sequence $\tightset{x_1,\dots,x_n}$ is unordered, and we provide no positional encodings (only learnable modality encodings \cite{lei2021less} to differentiate vision from language inputs), this operation is thus strictly \textit{atemporal}.%
\footnote{We include detailed experimental analysis and discussion of ATP atemporality (including \textit{relative} vs. \textit{absolute} encoder designs) in the supplement.}
These encodings $\tightset{s_1, \dots, s_n}$ are input to a final multilayer perceptron (MLP) to obtain logits for the final selection operation:
\begin{equation}
    MLP(\tightset{s_1, \dots, s_n}) \mapsto g \in \R^n.
\end{equation}

Our final selection operation ($S(g) \mapsto x_i$) is discrete: ATP must select a single embedding $x_i$. To ensure learnability, we consider two versions of our selector $S$ \textit{during training}, both operating on the logits $g$: the first employs a straight-through Gumbel-Softmax estimator~\cite{jang2016categorical}, the second applies softmax and ensures entropy decreases over time \cite{fanbuch2020rubiksnet}. In either case, at final test-time inference, the operation is made fully discrete; see supplement for details.%

\para{Training.} ATP is trained within the context of the overall video-language task framework, where the groundtruth answer or retrieval supervises the task loss, and gradients are backpropagated into the learnable ATP parameters. We re-iterate that no modifications are made to the frozen image-language encodings, and the final video-language task is performed directly on these frozen representations without any additional downstream learnable parameters. For both tasks, we optimize for the groundtruth similarity between the vision and language encodings. For video question answering, we consider a cross entropy loss over the answer set \cite{xiao2021next}, and for retrieval our loss is based on the standard InfoNCE contrastive loss \cite{radford2021learning}; see supplement for details.

\begin{figure}[t]
\begin{center}
\centerline{\includegraphics[width=0.95\linewidth]{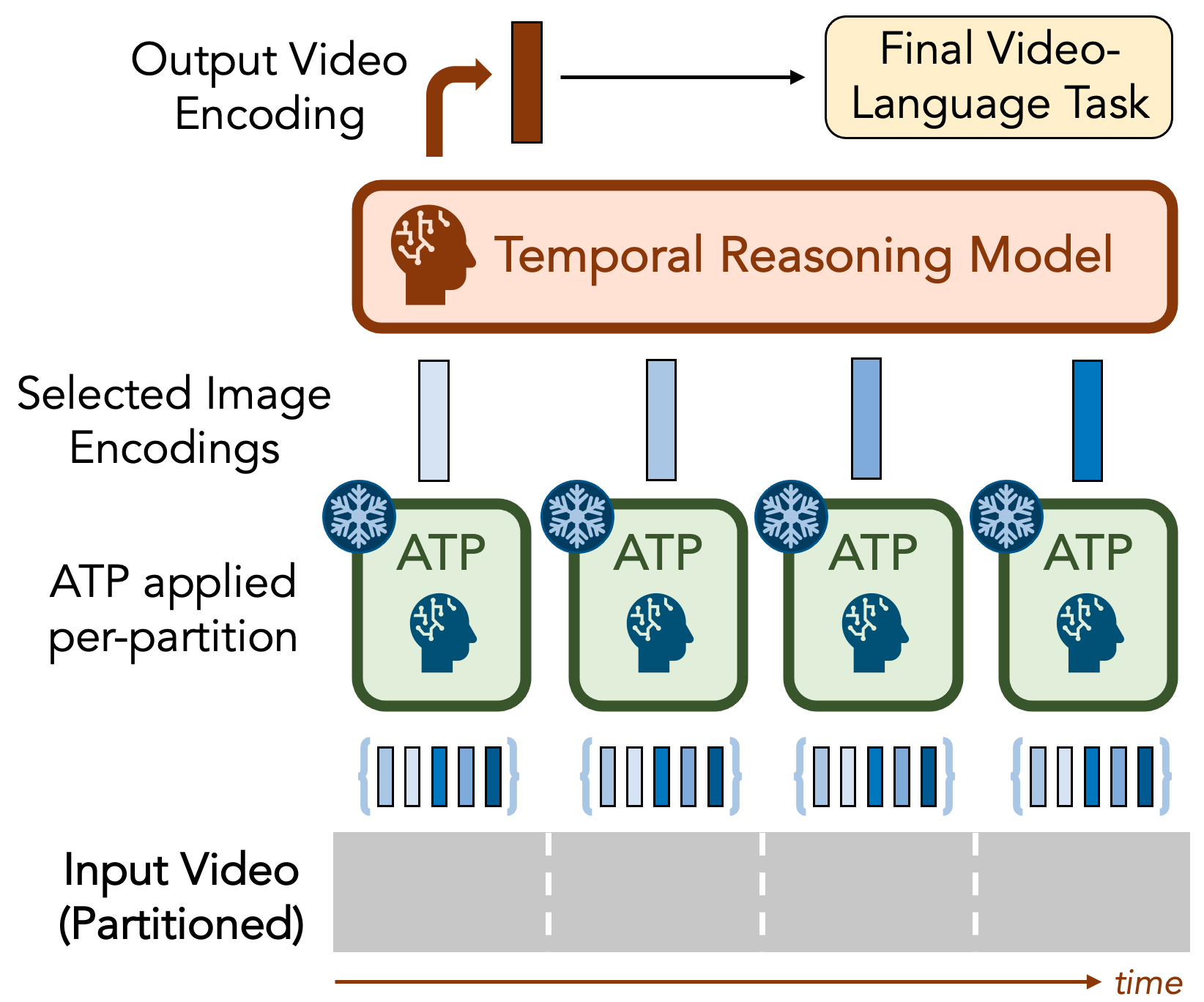}}
\vskip -0.1in
\caption{\textbf{Improving temporal modeling with ATP.} In our Section~\ref{sec:exp:next-qa} case study, we make use of learned single-embedding ATP selectors to improve temporal modeling. Intuitively, ATP learns to surface frames rich in single event-level information. Building upon this, we propose a simple approach to partition the original video and run (a now \textit{frozen}) ATP on each part. These per-partition selection outputs are then useful candidates for a separate downstream learnable model to perform temporal reasoning and output a video-level embedding for the final video-language task.
}
\label{fig:model-fig-temporal}
\end{center}
\vskip -0.2in
\end{figure}

\subsection{Improving Temporal Modeling with ATP}
\label{sec:ta:temporal-model}

In the final part of our experiments (Section~\ref{sec:experiments}), we additionally consider how our learned ATP embedding selector models (in Section~\ref{sec:ta:atp-model}) can improve downstream temporal models (Figure~\ref{fig:model-fig-temporal}). Intuitively, ATP learns to be an effective (language-conditional) event recognizer; building on this intuition, we propose a straightforward model that partitions the original video $V$ into $k$ partitions $V^{(1)}, \dots, V^{(k)}$ and runs (a learned, \textit{now frozen}) ATP model on each partition to obtain selected candidate embeddings $x_i^{(1)}, \dots,  x_j^{(k)}$ for the $k$ partitions. These per-partition outputs are then useful candidates for a separate, final learnable model $T$ that performs temporal reasoning and outputs a video-level embedding for the final video-language task. In Section~\ref{sec:exp:next-qa} experiments, this \textit{downstream} temporal model is a distinct transformer model, equipped to perform video-level reasoning \textit{on top of} ATP's output selections (for details, see supplement).

\section{Experiments}
\label{sec:experiments}

\subsection{Benchmark and Implementation Details}
\label{sec:exp:bench-implem}

\para{Benchmarks.} We consider three representative benchmarks for video question answering: NExT-QA \cite{xiao2021next}, VALUE-How2QA \cite{li2021value,li2020hero}, and MSR-VTT-MC \cite{xu2016msr}. We also examine the generality of our ATP model for text-to-video retrieval on DiDeMo \cite{hendricks2018localizing}, MSR-VTT \cite{xu2016msr}, and ActivityNet \cite{krishna2017dense}. For each benchmark, we follow standard protocols outlined by prior work \cite{lei2021less,xu2021videoclip,xiao2021next,li2021value} for dataset processing, metrics, and settings; see supplement for details and analysis. We choose these benchmarks specifically to provide a broad coverage of durations, source video domains (general activities, instructional, etc.), and designs. 

\para{Implementation.} We implement our ATP model with a few-layer, low-capacity transformer \cite{vaswani2017attention} in PyTorch \cite{paszke2019pytorch}, and train all models using the Adam \cite{kingma2014adam} optimizer. Main paper results here reported on ViT-B-32 (CLIP) inputs for consistency \cite{radford2018improving,dosovitskiy2020image,li2021value}.
See supplement%
\footnote{Please see project website for supplementary material and code release.}
for more.

\begin{figure}[t]
\begin{center}
\centerline{\includegraphics[width=0.95\linewidth]{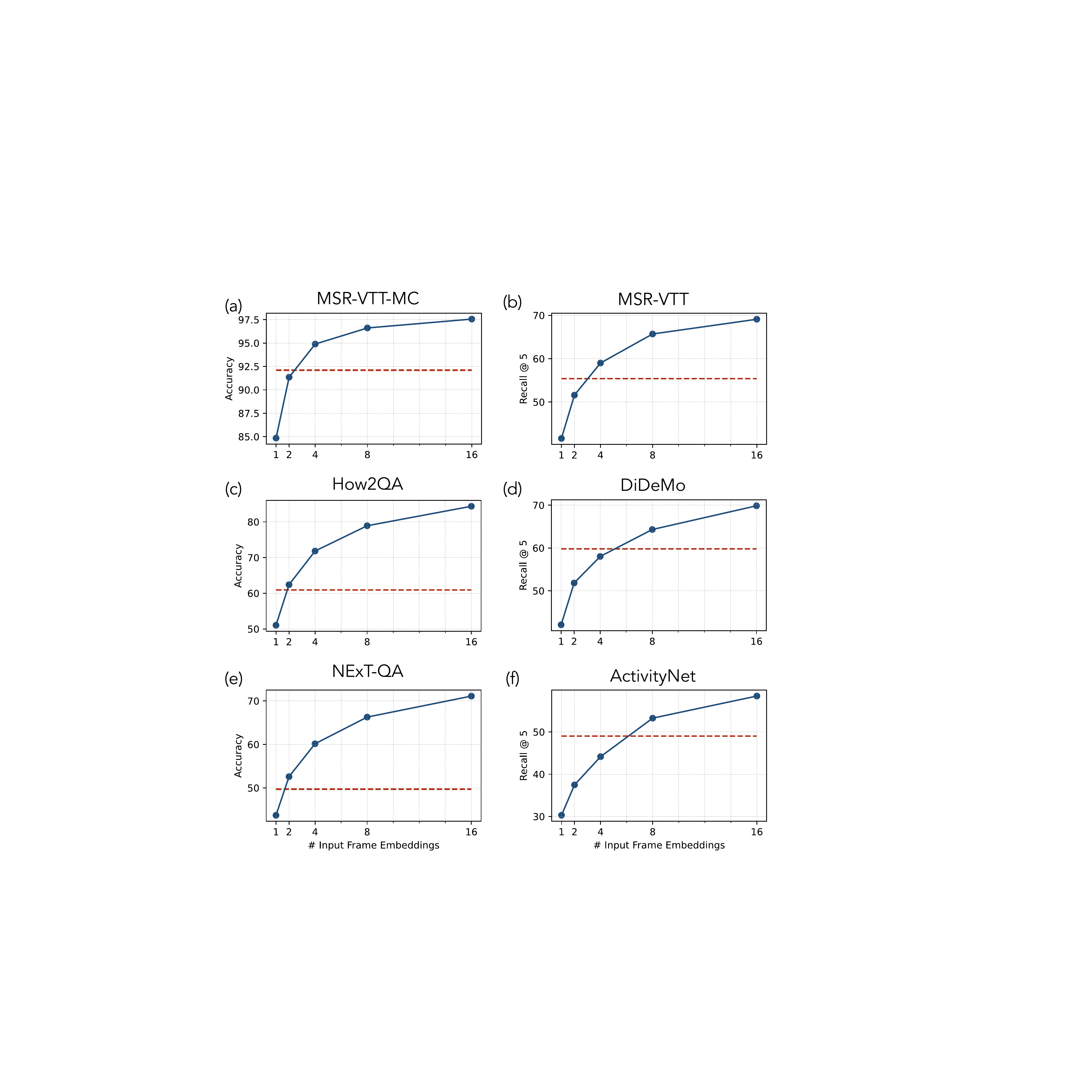}}
\vskip -0.1in
\caption{\textbf{Oracle upper bound analysis.} As a preliminary step, we analyze the performance upper bound of ATP under oracle conditions (with respect to the downstream task). Recall and accuracy (y-axis) averaged over multiple random samples of $n$ frames (x-axis). We observe that the upper bounds are competitive with state-of-the-art video models even when choosing one embedding from relatively few-frame samples. Dashed reference lines are state-of-the-art models (\textbf{(a,b)} \cite{xu2021videoclip}, \textbf{(c)} \cite{li2021value}, \textbf{(d)} \cite{bain2021frozen}, \textbf{(e)} \cite{xiao2021next}, \textbf{(f)} \cite{lei2021less}).
}
\label{fig:exp-oracle-analysis}
\end{center}
\vskip -0.2in
\end{figure}

\begin{figure*}[t]
\begin{center}
\centerline{\includegraphics[width=0.95\linewidth]{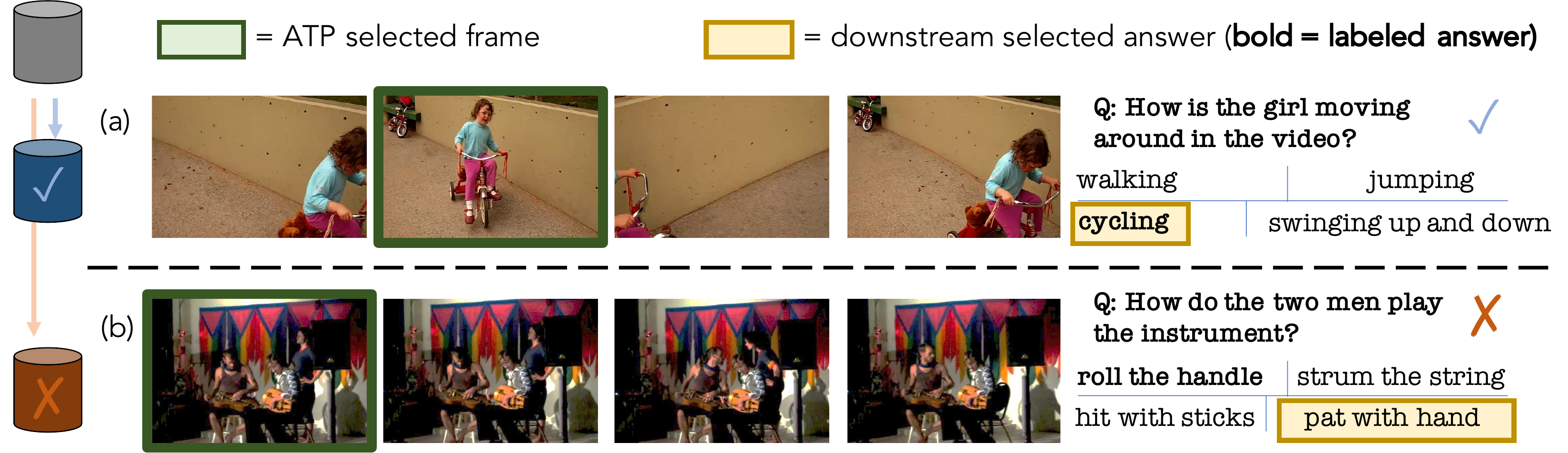}}
\vskip -0.1in
\caption{\textbf{ATP analysis (qualitative results).} We visualize example videos from the NExT-QA dataset \cite{xiao2021next}, along with the selections ATP made from a random sparse sample of frames. Both questions shown here are examples of ``causal-how'' questions in the dataset (shown with the top-4 answer options, for clarity). \textbf{(a)} We find that our ATP model can select informative frames for the downstream Video QA task, when possible, and that many questions initially intended to assess causal or temporal understanding can be answered from single-frame semantics. \textbf{(b)} Conversely, for (video, question) inputs that necessitate a deeper \textit{multi-frame} understanding of event relationships or dynamics, ATP's selected embedding is insufficient to answer the query. See Sec.~\ref{sec:exp:atp} (additional visuals and datasets in supplement).
}
\label{fig:exp-nextqa-qual}
\end{center}
\vskip -0.3in
\end{figure*}

\subsection{Analyzing Video-Language with ATP}
\label{sec:exp:atp}

\para{Preliminary (upper bound) analysis.} As a preliminary step, we first examine the performance of ATP under oracle conditions (with respect to the downstream video-language task) to establish a kind of upper bound for ATP on the set of benchmarks. In this analysis, we sample $n$ input frames from the video, varying $n$, and encode them with a pretrained model. In this oracle setting \textit{only}, ATP then selects a frozen embedding from this set that maximizes the downstream groundtruth accuracy on the video-language task. Note that in this analysis, the oracle empowered ATP is still bottlenecked by what the image-level representation is able to capture. We repeat this analysis for multiple samples (dependent on the video lengths), and report the average. As shown in Figure~\ref{fig:exp-oracle-analysis}, we observe that upper bound accuracies are competitive with state-of-the-art video models, even with relatively small $n$ sample sizes, suggesting the promise for analyzing these datasets with a learnable ATP.

\para{ATP analysis (video QA).} We apply a learnable ATP model to analyze a suite of standard video-language benchmarks. We first center our analysis discussion on video question-answering (video QA) benchmarks, since we find these benchmarks provide strong potential for deep multi-event understanding. Per Section~\ref{sec:exp:bench-implem}, we focus on three representative benchmarks for analysis: NExT-QA \cite{xiao2021next}, VALUE-How2QA \cite{li2021value,li2020hero}, and MSR-VTT-MC \cite{xu2016msr}. 
We re-iterate that our \textit{primary} goal with ATP is one of analysis: to better characterize these \textit{instantiations} of the video-language task.
In Tables \ref{tbl:vid-qa:msrvttmc}, \ref{tbl:vid-qa:how2qa-value}, and \ref{tbl:vid-qa:next-qa}, we report results for each benchmark.

\begin{table}[t]
\centering
\scalebox{0.83}{
\begin{tabular}{l c}
\textit{MSR-VTT-MC} & Accuracy  \\
\hline
ActBERT\cite{zhu2020actbert} & 85.7\\
ClipBERT\cite{lei2021less} & 88.2\\
MERLOT \cite{zellersluhessel2021merlot} & 90.9\\
VideoCLIP \cite{xu2021videoclip} & 92.1\\
\hline
    CLIP (single-frame) & 84.8 \\ 
    Ours (ATP; $1 \leftarrow 4 $)& 91.4 \\ 
    Ours (ATP; $1 \leftarrow 8 $)& \textbf{92.5} \\ 
    Ours (ATP; $1 \leftarrow 16 $)& \textbf{93.2} \\ 
\end{tabular}
}
\caption{\textbf{VideoQA on MSR-VTT-MC.} We find that our learned ATP model significantly outperforms prior work, indicating that this dataset can be largely addressed with image-level understanding. (1 $\leftarrow n$ means 1 embedding chosen from $n$ sampled.)}
\label{tbl:vid-qa:msrvttmc}
\end{table}

\begin{table}[bt]
\centering
\scalebox{0.83}{
\begin{tabular}{l c}
\textit{VALUE-How2QA} & Accuracy  \\
\hline
Random & 25.0 \\
HERO \cite{li2020hero} & 60.4\\
HERO+ \cite{li2020hero} & 60.9\\ 
\hline
	CLIP (single-frame) & 50.1 \\
	CLIP (mean pooling) & 55.7 \\ 
	Ours (ATP)& \textbf{65.1} \\ 
\end{tabular}
}
\caption{\textbf{VideoQA on VALUE-How2QA.} We observe strong performance over previous state-of-the-art baselines on instructional video data. HERO+ baseline here has the same preprocessing as our model, and all models leverage the same CLIP features (HERO baselines additionally leverage heavy motion features \cite{feichtenhofer2019slowfast,li2021value}).}
\label{tbl:vid-qa:how2qa-value}
\end{table}

On MSR-VTT-MC (Table~\ref{tbl:vid-qa:msrvttmc}), our learned ATP model outperforms recent state-of-the-art video-language models \cite{xu2021videoclip,zellersluhessel2021merlot,lei2021less}, when considering relatively few frames at inference and despite its substantial (single-frame) bottleneck constraints on model capacity, capability, and inputs. Critically, ATP substantially improves over standard atemporal baselines, including random single-frame and mean-pooling with CLIP \cite{radford2021learning}, offering a stronger bound.

On VALUE-How2QA (Table~\ref{tbl:vid-qa:how2qa-value}), we find that our learned ATP model offers significantly stronger accuracies than prior state-of-the-art models. Note that the HERO baselines here also use the same input CLIP embeddings, and no auxiliary text inputs, for fair comparison. One takeaway finding from our analysis of this benchmark was that counting questions, often designed to track state over the course of a video, were in fact often addressable by a single well-chosen frame that showed sufficient number of the items.

\begin{table}[bt]
\centering
\scalebox{0.83}{
    \begin{tabular}{l c c c c}
\textit{NExT-QA} & Acc  & Acc-D  & Acc-T  & Acc-C  \\
\hline
Random & 20.0 & 20.0 & 20.0 & 20.0 \\
\hline
\textsc{Main Dataset} & \multicolumn{4}{r}{(\textit{Section~\ref{sec:exp:atp}})}\\
	CLIP (single-frame) & 43.7 & 53.1 & 39.0 & 43.8 \\ 
	HGA \cite{jiang2020reasoning} & 49.7 & 59.3 & 50.7 & 46.3\\
	HGA \cite{jiang2020reasoning} + CLIP \cite{radford2021learning} & 50.4 & 59.3 & 52.1 & 46.8\\
    Ours (ATP) & 49.2  & 58.9 & 46.7 & 48.3 \\
\hline
    Ours (Temp[ATP])& \textbf{51.5} & 65.0 & 49.3 & 48.6 \\ 
    Ours (Temp[ATP] + ATP)& \textbf{54.3} & 66.8 & 50.2 & 53.1 \\ 
\hline
\hline
\textsc{ATP$_{hard}$-Subset} & \multicolumn{4}{r}{(\textit{Section~\ref{sec:exp:next-qa}})}\\
    Ours (ATP) & 20.2 & 23.9 & 22.6 & 19.6 \\ 
    Ours (Temporal[ATP])& 38.8 & 46.8 & 36.5 & 38.4 \\ 
	HGA \cite{jiang2020reasoning} & 44.1 & 51.2 & 45.3 & 43.3 \\
\end{tabular}
}
\caption{\textbf{VideoQA on NExT-QA.} We report accuracies on the overall main dataset and descriptive (D), temporal (T), and causal (C) splits. See Section~\ref{sec:exp:next-qa} for details on the ``Temp[ATP]'' and ``Temp[ATP] + ATP'' models, and details on our ATP$_{hard}$ subset.}
\vspace{-.5em}
\label{tbl:vid-qa:next-qa}
\end{table}

Finally, on NExT-QA (Table~\ref{tbl:vid-qa:next-qa}), we find that even this recent benchmark, which is explicitly designed for temporal and causal understanding, can have a non-trivial subset of questions answerable by simple single-frame event recognition. In Figure~\ref{fig:exp-nextqa-qual}, we show two different ``causal-how'' questions, which aim to assess both causality and dynamics. In the case of  Figure~\ref{fig:exp-nextqa-qual}(a) specifically, we observe that as long as the ATP model is able to select the informative frame with a clear depiction of the child on the cycle, the answer is readily apparent without deep video-level understanding. Quantitatively, our ATP model provides a stronger bound than standard image-level baselines; we also augment the the HGA baseline with CLIP features for fair comparison.

\para{ATP analysis (retrieval).} We also apply our learnable ATP model on standard retrieval benchmarks: DiDeMo \cite{hendricks2018localizing}, MSR-VTT \cite{xu2016msr}, and ActivityNet \cite{krishna2017dense}. In Table~\ref{tbl:vid-lang:all}, we observe that our technique generalizes well to other discriminative video-language settings, establishing stronger bounds on image-centric performance and showing competitive accuracies with recent state-of-the-art methods. Our ATP model's performance on \textit{paragraph} retrieval settings, like ActivityNet, highlights an area of improvement for image-bottleneck understanding: because paragraphs describe multiple dense events in long videos, it can be difficult to use a single frame embedding to capture this description well. We provide an extended discussion of prior work comparisons, limitations, and potential future directions for text-to-video retrieval as part of our supplement.

\begin{table}[t]

\centering
\setlength\tabcolsep{0.1pt}
\scalebox{.83}{
    \begin{tabular}{l |c c | c c | c c}
  & \multicolumn{2}{|c|}{\textit{MSR-VTT}} & \multicolumn{2}{c|}{\textit{DiDeMo}} & \multicolumn{2}{c}{\textit{ActivityNet}} \\
  & ~R@1~  & ~R@5~  & ~R@1~  & ~R@5~  & ~R@1~  & ~R@5~  \\
\hline
     Support Set\cite{patrick2021supportset} & 30.1 & 58.5 & - & - & 29.2 & 61.6\\
     VideoCLIP \cite{xu2021videoclip} & 30.9 & 55.4 & 16.6$^*$ & 46.9$^*$ & - & - \\
     ClipBERT \cite{lei2021less} & 22.0 & 46.8 & 20.4 & 48.0 & 21.3 & 49.0\\
	\hline   
    CLIP (single-frame)~ & 21.6 & 44.6 & 20.2 & 42.5 & 12.5 & 30.3 \\ 
    Ours (ATP)& 27.8 & 49.8 & 26.1 & 50.5 & 17.7 & 41.8\\ 
    \end{tabular}
}
\caption{\textbf{Video-language (text-to-video) retrieval.} We show that our ATP analysis technique generalizes beyond video question answering settings. ($^*$ indicates zero-shot reported by prior work; see supplement for a more complete prior work comparisons table.)}
\label{tbl:vid-lang:all}
\end{table}

\subsection{Improving Dataset and Model Design with ATP}
\label{sec:exp:next-qa}

Finally, we consider how to leverage our ATP model and its insights to improve both dataset and model design. For this section, we choose to focus on the NExT-QA benchmark \cite{xiao2021next} as a case study, since it is a key recent effort towards improving the field's focus on causal and temporal understanding in video-language tasks.

\para{Improving dataset design with ATP.} From our initial analysis of the NExT-QA benchmark in Section~\ref{sec:exp:atp}, we found that ATP provides a surprising degree of accuracy on causal and temporal questions, despite its strong image-centric bottleneck. Because ATP provides a stronger bound on the capability of image-level understanding for these questions, it can help better disentangle questions that necessitate full video-level understanding (such questions will be largely unanswerable for the ATP model) from ones that do not. 

\begin{figure}[t]
\begin{center}
\centerline{\includegraphics[width=\linewidth]{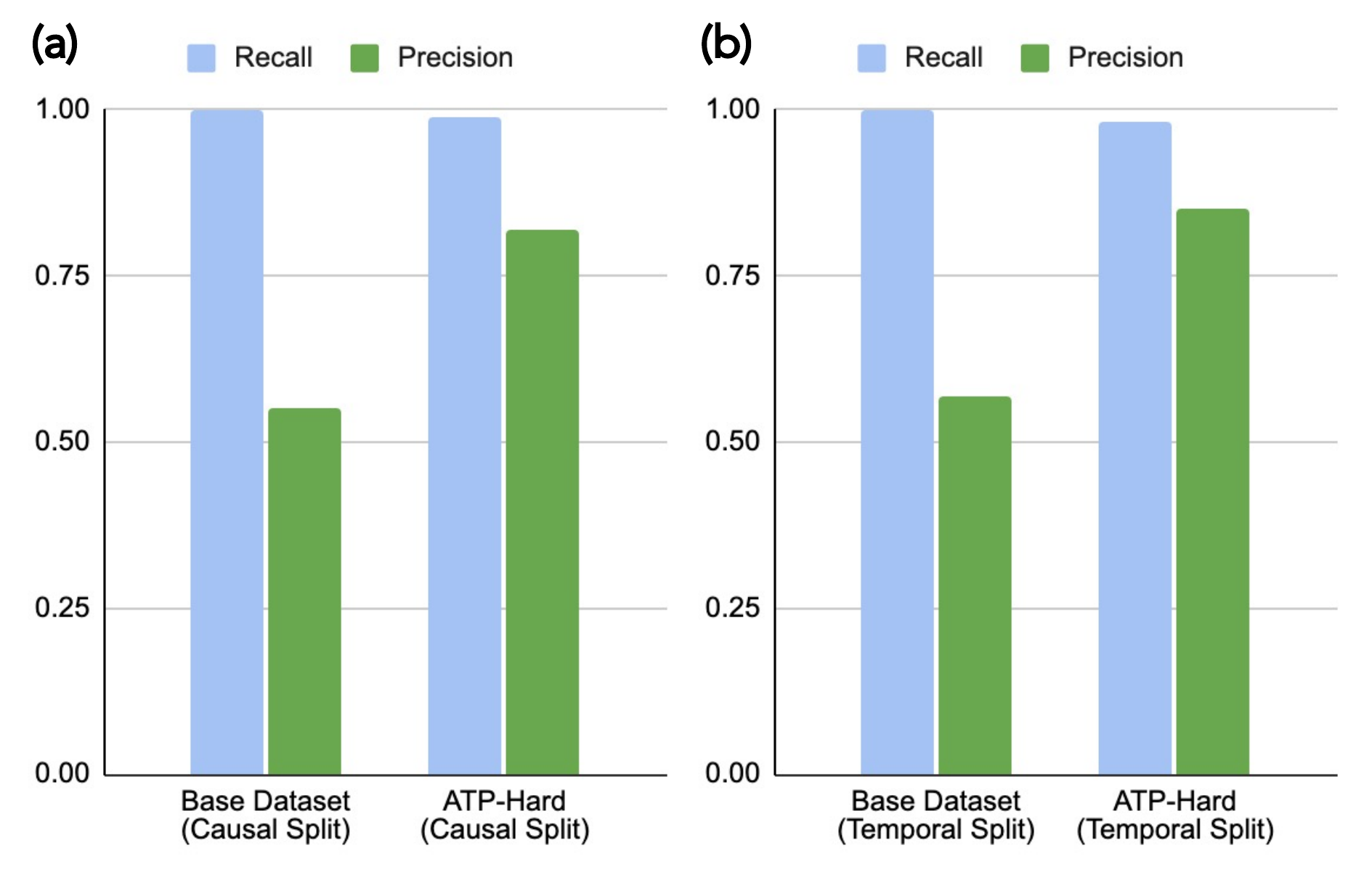}}
\vskip -0.1in
\caption{\textbf{Improving dataset design with ATP.} We analyze the original NExT-QA \cite{xiao2021next} benchmark, and find that our ATP models (denoted ATP$_{hard}$) are able to better disentangle input (video, question) pairs that \textit{truly} necessitate video-level understanding compared with the original dataset, on both \textbf{(a)} casual and \textbf{(b)} temporal splits. This indicates promise for leveraging ATP in-the-loop for future dataset designs. See Section~\ref{sec:exp:next-qa} for analysis details.
}
\label{fig:exp-nextqa-disentanglement}
\end{center}
\vskip -0.3in
\end{figure}

We accomplish this by considering an ensemble of ATP models on the dataset, and leveraging their confidences and agreement to determine a subset of ATP$_{hard}$ questions. We determine any heuristics through k-fold cross validation on the \textit{training} set. In parallel, we manually annotate a subset of the validation set for (video, question) pairs that predicate video-level understanding (see supplement for procedure details and limitations of our ATP technique). The results of our final analysis are shown in Figure~\ref{fig:exp-nextqa-disentanglement}. We find that our ATP based technique maintains the recall of the video-level understanding questions on both the causal and temporal dataset splits, while simultaneously improving upon their precision (by filtering out ``easy'' questions).

Furthermore, we can also show how this ATP$_{hard}$ subset better benchmarks progress on video-level causal and temporal understanding (in Table~\ref{tbl:vid-qa:next-qa}) that may have been otherwise obscured. While ATP nearly matches the other models on the main dataset due to the inclusion  of ``easier'' questions, this harder subset reveals a substantial gap relative to the state-of-the-art temporal reasoning model. 

Together, these results suggest ATP in-the-loop can be an effective tool during future dataset design and creation.

\para{Improving model design with ATP.} As described in Section~\ref{sec:ta:temporal-model}, we can leverage ATP to provide candidate frame embeddings for a downstream temporal model. As a first step towards improving temporal modeling (and efficiency), we introduce this model (denoted Temp[ATP] in Table~\ref{tbl:vid-qa:next-qa}) and benchmark it on the NExT-QA dataset. This model achieves a new state-of-the-art accuracy, outperforming the HGA (and HGA + CLIP) baselines on the main NExT-QA dataset, while operating at significantly reduced processing cost due to ATP (see supplement for efficiency discussion).

We make two additional observations: first, on the ATP$_{hard}$ subset, we find that this temporal model recovers much of the performance gap between ATP and the HGA model (we attribute the remaining gap to HGA's incorporation of additional motion features, which can aid in addressing some challenging dynamics questions), further verifying the potential dataset design contribution of ATP. Second, we observe that the aggregated confidence scores of the ATP ensemble provides a clear disentanglement signal on hard vs. easy problems, without access to groundtruth. 
Setting a heuristic threshold with k-fold cross validation on training, we use this signal to smartly ensemble ATP and Temp[ATP] further. For questions ATP can address, we do not need to consider additional (potentially noisy) frames, and we can skip the full temporal model. For ones ATP is less confident, the temporal model is run. This ensemble (denoted Temp[ATP] + ATP in Table~\ref{tbl:vid-qa:next-qa}) achieves a significant further accuracy and efficiency increase on NExT-QA.

%

\section{Conclusion}

In this work, we revisit a fundamental question of video understanding (\textit{what makes a video task uniquely suited for videos, beyond what can be understood from a single image?}), building beyond prior analyses in action recognition towards video-language settings with more complex events.
First, we propose an atemporal probe (ATP) model to provide a stronger bound on how much of video-language understanding can be addressed from image-language understanding only.
Second, we use ATP to characterize both the limitations and potential of current video-language benchmarks for video question answering and video-language retrieval. Surprisingly, we find that single frame understanding can often achieve strong performance, even in settings intended for complex multi-frame event understanding and compared with recent large-scale video models.
Third, we show how ATP can be leveraged to improve designs for both video-language datasets (disentangling unintentional atemporal biases) and video-level models (improving efficiency and accuracy).
Going forward, we envision ATP as joining a broader, standard toolkit for video-language researchers and practitioners, revealing insights into complementary, \textit{video-specific} sources of bias in multimodal video settings.

\para{Broader Impacts and Limitations.} We provide a detailed discussion of limitations and implications for broader impacts of our proposed ATP and analysis in our supplement.

\para{Acknowledgements.} This work is supported in part by Toyota Research Institute (TRI), the Stanford Institute for Human-Centered AI (HAI), Samsung, Salesforce, and an NDSEG Fellowship (for S.B.). This article reflects the authors’ opinions and conclusions, and not any other entity. We also thank our colleagues in the Stanford Vision and Learning Lab (including Jim Fan, Ajay Mandlekar, Andrey Kurenkov, Sanjana Srivastava, Boxiao Pan, Ehsan Adeli) and the Stanford NLP Group (including Alex Tamkin, Percy Liang, Siddharth Karamcheti) for valuable discussions and support. We also thank Linjie Li and the authors of benchmarks examined here for their help towards enabling more direct comparisons with prior work, as well as our anonymous reviewers for helpful suggestions and feedback.

%

{\small
\bibliographystyle{ieee_fullname}
\bibliography{0_main_arxiv}

\begin{thebibliography}{10}\itemsep=-1pt

\bibitem{antol2015vqa}
Stanislaw Antol, Aishwarya Agrawal, Jiasen Lu, Margaret Mitchell, Dhruv Batra,
  C Lawrence~Zitnick, and Devi Parikh.
\newblock Vqa: Visual question answering.
\newblock In {\em ICCV}, 2015.

\bibitem{bain2021frozen}
Max Bain, Arsha Nagrani, G{\"u}l Varol, and Andrew Zisserman.
\newblock Frozen in time: A joint video and image encoder for end-to-end
  retrieval.
\newblock {\em arXiv preprint arXiv:2104.00650}, 2021.

\bibitem{carreira2017quo}
Joao Carreira and Andrew Zisserman.
\newblock Quo vadis, action recognition? a new model and the kinetics dataset.
\newblock In {\em proceedings of the IEEE Conference on Computer Vision and
  Pattern Recognition}, pages 6299--6308, 2017.

\bibitem{chen2015microsoft}
Xinlei Chen, Hao Fang, Tsung-Yi Lin, Ramakrishna Vedantam, Saurabh Gupta, Piotr
  Doll{\'a}r, and C~Lawrence Zitnick.
\newblock Microsoft coco captions: Data collection and evaluation server.
\newblock {\em arXiv}, 2015.

\bibitem{devlin2018bert}
Jacob Devlin, Ming-Wei Chang, Kenton Lee, and Kristina Toutanova.
\newblock Bert: Pre-training of deep bidirectional transformers for language
  understanding.
\newblock In {\em NAACL}, 2019.

\bibitem{diba2020large}
Ali Diba, Mohsen Fayyaz, Vivek Sharma, Manohar Paluri, J{\"u}rgen Gall, Rainer
  Stiefelhagen, and Luc Van~Gool.
\newblock Large scale holistic video understanding.
\newblock In {\em European Conference on Computer Vision}, pages 593--610.
  Springer, 2020.

\bibitem{dosovitskiy2020image}
Alexey Dosovitskiy, Lucas Beyer, Alexander Kolesnikov, Dirk Weissenborn,
  Xiaohua Zhai, Thomas Unterthiner, Mostafa Dehghani, Matthias Minderer, Georg
  Heigold, Sylvain Gelly, et~al.
\newblock An image is worth 16x16 words: Transformers for image recognition at
  scale.
\newblock {\em ICLR}, 2021.

\bibitem{engin2021hidden}
Deniz Engin, Fran{\c{c}}ois Schnitzler, Ngoc~QK Duong, and Yannis Avrithis.
\newblock On the hidden treasure of dialog in video question answering.
\newblock In {\em Proceedings of the IEEE/CVF International Conference on
  Computer Vision}, pages 2064--2073, 2021.

\bibitem{fanbuch2020rubiksnet}
Linxi Fan*, Shyamal Buch*, Guanzhi Wang, Ryan Cao, Yuke Zhu, Juan~Carlos
  Niebles, and Li Fei-Fei.
\newblock {RubiksNet: Learnable 3D-Shift for Efficient Video Action
  Recognition}.
\newblock In {\em European Conference on Computer Vision}, pages 505--521.
  Springer, 2020.

\bibitem{feichtenhofer2020x3d}
Christoph Feichtenhofer.
\newblock X3d: Expanding architectures for efficient video recognition.
\newblock In {\em CVPR}, 2020.

\bibitem{feichtenhofer2019slowfast}
Christoph Feichtenhofer, Haoqi Fan, Jitendra Malik, and Kaiming He.
\newblock Slowfast networks for video recognition.
\newblock In {\em ICCV}, 2019.

\bibitem{girdhar2019cater}
Rohit Girdhar and Deva Ramanan.
\newblock Cater: A diagnostic dataset for compositional actions and temporal
  reasoning.
\newblock {\em arXiv preprint arXiv:1910.04744}, 2019.

\bibitem{goyal2017something}
Raghav Goyal, Samira Ebrahimi~Kahou, Vincent Michalski, Joanna Materzynska,
  Susanne Westphal, Heuna Kim, Valentin Haenel, Ingo Fruend, Peter Yianilos,
  Moritz Mueller-Freitag, et~al.
\newblock The" something something" video database for learning and evaluating
  visual common sense.
\newblock In {\em Proceedings of the IEEE international conference on computer
  vision}, pages 5842--5850, 2017.

\bibitem{goyal2017making}
Yash Goyal, Tejas Khot, Douglas Summers-Stay, Dhruv Batra, and Devi Parikh.
\newblock Making the v in vqa matter: Elevating the role of image understanding
  in visual question answering.
\newblock In {\em CVPR}, 2017.

\bibitem{he2016deep}
Kaiming He, Xiangyu Zhang, Shaoqing Ren, and Jian Sun.
\newblock Deep residual learning for image recognition.
\newblock In {\em Proceedings of the IEEE conference on computer vision and
  pattern recognition}, pages 770--778, 2016.

\bibitem{hendricks2018localizing}
Lisa~Anne Hendricks, Oliver Wang, Eli Shechtman, Josef Sivic, Trevor Darrell,
  and Bryan Russell.
\newblock Localizing moments in video with temporal language.
\newblock In {\em Empirical Methods in Natural Language Processing (EMNLP)},
  2018.

\bibitem{huang2018what}
De-An Huang, Vignesh Ramanathan, Dhruv Mahajan, Lorenzo Torresani, Manohar
  Paluri, Li Fei-Fei, and Juan~Carlos Niebles.
\newblock What makes a video a video: Analyzing temporal information in video
  understanding models and datasets.
\newblock In {\em Proceedings of the IEEE Conference on Computer Vision and
  Pattern Recognition}, pages 7366--7375, 2018.

\bibitem{jang2016categorical}
Eric Jang, Shixiang Gu, and Ben Poole.
\newblock Categorical reparameterization with gumbel-softmax.
\newblock {\em arXiv preprint arXiv:1611.01144}, 2016.

\bibitem{jang2017tgif}
Yunseok Jang, Yale Song, Youngjae Yu, Youngjin Kim, and Gunhee Kim.
\newblock Tgif-qa: Toward spatio-temporal reasoning in visual question
  answering.
\newblock In {\em CVPR}, 2017.

\bibitem{jia2021scaling}
Chao Jia, Yinfei Yang, Ye Xia, Yi-Ting Chen, Zarana Parekh, Hieu Pham, Quoc~V
  Le, Yunhsuan Sung, Zhen Li, and Tom Duerig.
\newblock Scaling up visual and vision-language representation learning with
  noisy text supervision.
\newblock {\em ICML}, 2021.

\bibitem{jiang2020reasoning}
Pin Jiang and Yahong Han.
\newblock Reasoning with heterogeneous graph alignment for video question
  answering.
\newblock In {\em Proceedings of the AAAI Conference on Artificial
  Intelligence}, 2020.

\bibitem{kingma2014adam}
Diederik~P Kingma and Jimmy Ba.
\newblock Adam: A method for stochastic optimization.
\newblock {\em arXiv preprint arXiv:1412.6980}, 2014.

\bibitem{krishna2017dense}
Ranjay Krishna, Kenji Hata, Frederic Ren, Li Fei-Fei, and Juan~Carlos Niebles.
\newblock Dense-captioning events in videos.
\newblock In {\em International Conference on Computer Vision (ICCV)}, 2017.

\bibitem{kuehne2011hmdb}
Hildegard Kuehne, Hueihan Jhuang, Est{\'\i}baliz Garrote, Tomaso Poggio, and
  Thomas Serre.
\newblock Hmdb: a large video database for human motion recognition.
\newblock In {\em 2011 International conference on computer vision}, pages
  2556--2563. IEEE, 2011.

\bibitem{lee2019set}
Juho Lee, Yoonho Lee, Jungtaek Kim, Adam Kosiorek, Seungjin Choi, and Yee~Whye
  Teh.
\newblock Set transformer: A framework for attention-based
  permutation-invariant neural networks.
\newblock In {\em International Conference on Machine Learning}, pages
  3744--3753. PMLR, 2019.

\bibitem{lei2021less}
Jie Lei, Linjie Li, Luowei Zhou, Zhe Gan, Tamara~L. Berg, Mohit Bansal, and
  Jingjing Liu.
\newblock Less is more: Clipbert for video-and-language learningvia sparse
  sampling.
\newblock In {\em CVPR}, 2021.

\bibitem{lei2018tvqa}
Jie Lei, Licheng Yu, Mohit Bansal, and Tamara~L Berg.
\newblock Tvqa: Localized, compositional video question answering.
\newblock In {\em EMNLP}, 2018.

\bibitem{lei2020vlep}
Jie Lei, Licheng Yu, Tamara~L Berg, and Mohit Bansal.
\newblock What is more likely to happen next? video-and-language future event
  prediction.
\newblock In {\em EMNLP}, 2020.

\bibitem{li2020hero}
Linjie Li, Yen-Chun Chen, Yu Cheng, Zhe Gan, Licheng Yu, and Jingjing Liu.
\newblock {HERO}: Hierarchical encoder for {V}ideo+{L}anguage
  omni-representation pre-training.
\newblock In {\em Proceedings of the 2020 Conference on Empirical Methods in
  Natural Language Processing (EMNLP)}, pages 2046--2065, Online, Nov. 2020.
  Association for Computational Linguistics.

\bibitem{li2021value}
Linjie Li, Jie Lei, Zhe Gan, Licheng Yu, Yen-Chun Chen, Rohit Pillai, Yu Cheng,
  Luowei Zhou, Xin~Eric Wang, William~Yang Wang, et~al.
\newblock Value: A multi-task benchmark for video-and-language understanding
  evaluation.
\newblock {\em NeurIPS (Benchmarks and Datasets Track)}, 2021.

\bibitem{lin2019tsm}
Ji Lin, Chuang Gan, and Song Han.
\newblock Tsm: Temporal shift module for efficient video understanding.
\newblock In {\em Proceedings of the IEEE/CVF International Conference on
  Computer Vision}, pages 7083--7093, 2019.

\bibitem{liu2021no}
Xin Liu, Silvia~L Pintea, Fatemeh~Karimi Nejadasl, Olaf Booij, and Jan~C van
  Gemert.
\newblock No frame left behind: Full video action recognition.
\newblock In {\em Proceedings of the IEEE/CVF Conference on Computer Vision and
  Pattern Recognition}, pages 14892--14901, 2021.

\bibitem{liu2019roberta}
Yinhan Liu, Myle Ott, Naman Goyal, Jingfei Du, Mandar Joshi, Danqi Chen, Omer
  Levy, Mike Lewis, Luke Zettlemoyer, and Veselin Stoyanov.
\newblock Roberta: A robustly optimized bert pretraining approach.
\newblock {\em arXiv preprint arXiv:1907.11692}, 2019.

\bibitem{lu2021metadata}
Mandy Lu, Qingyu Zhao, Jiequan Zhang, Kilian~M Pohl, Li Fei-Fei, Juan~Carlos
  Niebles, and Ehsan Adeli.
\newblock Metadata normalization.
\newblock In {\em Proceedings of the IEEE/CVF Conference on Computer Vision and
  Pattern Recognition}, pages 10917--10927, 2021.

\bibitem{miech2020end}
Antoine Miech, Jean-Baptiste Alayrac, Lucas Smaira, Ivan Laptev, Josef Sivic,
  and Andrew Zisserman.
\newblock End-to-end learning of visual representations from uncurated
  instructional videos.
\newblock In {\em Proceedings of the IEEE/CVF Conference on Computer Vision and
  Pattern Recognition}, pages 9879--9889, 2020.

\bibitem{miech2019howto100m}
Antoine Miech, Dimitri Zhukov, Jean-Baptiste Alayrac, Makarand Tapaswi, Ivan
  Laptev, and Josef Sivic.
\newblock Howto100m: Learning a text-video embedding by watching hundred
  million narrated video clips.
\newblock In {\em Proceedings of the IEEE international conference on computer
  vision}, pages 2630--2640, 2019.

\bibitem{paszke2019pytorch}
Adam Paszke, Sam Gross, Francisco Massa, Adam Lerer, James Bradbury, Gregory
  Chanan, Trevor Killeen, Zeming Lin, Natalia Gimelshein, Luca Antiga, Alban
  Desmaison, Andreas Kopf, Edward Yang, Zachary DeVito, Martin Raison, Alykhan
  Tejani, Sasank Chilamkurthy, Benoit Steiner, Lu Fang, Junjie Bai, and Soumith
  Chintala.
\newblock Pytorch: An imperative style, high-performance deep learning library.
\newblock In {\em NeurIPS}, 2019.

\bibitem{patrick2021supportset}
Mandela Patrick, Po-Yao Huang, Yuki Asano, Florian Metze, Alexander~G
  Hauptmann, Joao~F. Henriques, and Andrea Vedaldi.
\newblock Support-set bottlenecks for video-text representation learning.
\newblock In {\em International Conference on Learning Representations}, 2021.

\bibitem{pillutla2021information}
Krishna Pillutla, Swabha Swayamdipta, Rowan Zellers, John Thickstun, Sean
  Welleck, Yejin Choi, and Zaid Harchaoui.
\newblock An information divergence measure between neural text and human text.
\newblock {\em NeurIPS}, 2021.

\bibitem{qiu2017learning}
Zhaofan Qiu, Ting Yao, and Tao Mei.
\newblock Learning spatio-temporal representation with pseudo-3d residual
  networks.
\newblock In {\em CVPR}, 2017.

\bibitem{radford2021learning}
Alec Radford, Jong~Wook Kim, Chris Hallacy, Aditya Ramesh, Gabriel Goh,
  Sandhini Agarwal, Girish Sastry, Amanda Askell, Pamela Mishkin, Jack Clark,
  et~al.
\newblock Learning transferable visual models from natural language
  supervision.
\newblock {\em arXiv preprint arXiv:2103.00020}, 2021.

\bibitem{radford2018improving}
Alec Radford, Karthik Narasimhan, Tim Salimans, and Ilya Sutskever.
\newblock Improving language understanding by generative pre-training, 2018.

\bibitem{rohrbach2015dataset}
Anna Rohrbach, Marcus Rohrbach, Niket Tandon, and Bernt Schiele.
\newblock A dataset for movie description.
\newblock In {\em CVPR}, 2015.

\bibitem{schindler2008action}
Konrad Schindler and Luc Van~Gool.
\newblock Action snippets: How many frames does human action recognition
  require?
\newblock In {\em 2008 IEEE Conference on Computer Vision and Pattern
  Recognition}, pages 1--8. IEEE, 2008.

\bibitem{shen2021much}
Sheng Shen, Liunian~Harold Li, Hao Tan, Mohit Bansal, Anna Rohrbach, Kai-Wei
  Chang, Zhewei Yao, and Kurt Keutzer.
\newblock How much can clip benefit vision-and-language tasks?
\newblock {\em arXiv preprint arXiv:2107.06383}, 2021.

\bibitem{simonyan2014two}
Karen Simonyan and Andrew Zisserman.
\newblock Two-stream convolutional networks for action recognition in videos.
\newblock In {\em NeurIPS}, 2014.

\bibitem{smaira2020short}
Lucas Smaira, Jo{\~a}o Carreira, Eric Noland, Ellen Clancy, Amy Wu, and Andrew
  Zisserman.
\newblock A short note on the kinetics-700-2020 human action dataset.
\newblock {\em arXiv preprint arXiv:2010.10864}, 2020.

\bibitem{soomro2012ucf101}
Khurram Soomro, Amir~Roshan Zamir, and Mubarak Shah.
\newblock Ucf101: A dataset of 101 human actions classes from videos in the
  wild.
\newblock {\em arXiv preprint arXiv:1212.0402}, 2012.

\bibitem{sun2019videobert}
Chen Sun, Austin Myers, Carl Vondrick, Kevin Murphy, and Cordelia Schmid.
\newblock Videobert: A joint model for video and language representation
  learning.
\newblock In {\em Proceedings of the IEEE International Conference on Computer
  Vision}, pages 7464--7473, 2019.

\bibitem{vaswani2017attention}
Ashish Vaswani, Noam Shazeer, Niki Parmar, Jakob Uszkoreit, Llion Jones,
  Aidan~N Gomez, {\L}ukasz Kaiser, and Illia Polosukhin.
\newblock Attention is all you need.
\newblock In {\em Advances in neural information processing systems}, pages
  5998--6008, 2017.

\bibitem{wang2016temporal}
Limin Wang, Yuanjun Xiong, Zhe Wang, Yu Qiao, Dahua Lin, Xiaoou Tang, and Luc
  Van~Gool.
\newblock Temporal segment networks: Towards good practices for deep action
  recognition.
\newblock In {\em European conference on computer vision}, pages 20--36.
  Springer, 2016.

\bibitem{lfb2019}
Chao-Yuan Wu, Christoph Feichtenhofer, Haoqi Fan, Kaiming He, Philipp
  Kr\"{a}henb\"{u}hl, and Ross Girshick.
\newblock {Long-Term Feature Banks for Detailed Video Understanding}.
\newblock In {\em {CVPR}}, 2019.

\bibitem{wu2019adaframe}
Zuxuan Wu, Caiming Xiong, Chih-Yao Ma, Richard Socher, and Larry~S Davis.
\newblock Adaframe: Adaptive frame selection for fast video recognition.
\newblock In {\em CVPR}, 2019.

\bibitem{xiao2021next}
Junbin Xiao, Xindi Shang, Angela Yao, and Tat-Seng Chua.
\newblock {NExT-QA:} next phase of question-answering to explaining temporal
  actions.
\newblock In {\em Proceedings of the IEEE/CVF Conference on Computer Vision and
  Pattern Recognition (CVPR)}, pages 9777--9786, June 2021.

\bibitem{xu2021videoclip}
Hu Xu, Gargi Ghosh, Po-Yao Huang, Dmytro Okhonko, Armen Aghajanyan, Florian
  Metze, Luke Zettlemoyer, and Christoph Feichtenhofer.
\newblock {VideoCLIP}: Contrastive pre-training for zero-shot video-text
  understanding.
\newblock In {\em Proceedings of the 2021 Conference on Empirical Methods in
  Natural Language Processing (EMNLP)}, Online, Nov. 2021. Association for
  Computational Linguistics.

\bibitem{xu2016msr}
Jun Xu, Tao Mei, Ting Yao, and Yong Rui.
\newblock Msr-vtt: A large video description dataset for bridging video and
  language.
\newblock In {\em Proceedings of the IEEE conference on computer vision and
  pattern recognition}, pages 5288--5296, 2016.

\bibitem{yeung2016end}
Serena Yeung, Olga Russakovsky, Greg Mori, and Li Fei-Fei.
\newblock End-to-end learning of action detection from frame glimpses in
  videos.
\newblock In {\em Proceedings of the IEEE conference on computer vision and
  pattern recognition}, pages 2678--2687, 2016.

\bibitem{yi2019clevrer}
Kexin Yi, Chuang Gan, Yunzhu Li, Pushmeet Kohli, Jiajun Wu, Antonio Torralba,
  and Joshua~B Tenenbaum.
\newblock Clevrer: Collision events for video representation and reasoning.
\newblock {\em arXiv preprint arXiv:1910.01442}, 2019.

\bibitem{yu2019activityqa}
Zhou Yu, Dejing Xu, Jun Yu, Ting Yu, Zhou Zhao, Yueting Zhuang, and Dacheng
  Tao.
\newblock Activitynet-qa: A dataset for understanding complex web videos via
  question answering.
\newblock In {\em AAAI}, pages 9127--9134, 2019.

\bibitem{zellersluhessel2021merlot}
Rowan Zellers, Ximing Lu, Jack Hessel, Youngjae Yu, Jae~Sung Park, Jize Cao,
  Ali Farhadi, and Yejin Choi.
\newblock Merlot: Multimodal neural script knowledge models.
\newblock {\em {NeurIPS}}, 2021.

\bibitem{zhou2018temporal}
Bolei Zhou, Alex Andonian, Aude Oliva, and Antonio Torralba.
\newblock Temporal relational reasoning in videos.
\newblock In {\em Proceedings of the European Conference on Computer Vision
  (ECCV)}, pages 803--818, 2018.

\bibitem{zhou2017towards}
Luowei Zhou, Chenliang Xu, and Jason~J Corso.
\newblock Towards automatic learning of procedures from web instructional
  videos.
\newblock In {\em AAAI}, 2018.

\bibitem{zhu2020actbert}
Linchao Zhu and Yi Yang.
\newblock Actbert: Learning global-local video-text representations.
\newblock In {\em Proceedings of the IEEE/CVF Conference on Computer Vision and
  Pattern Recognition}, pages 8746--8755, 2020.

\end{thebibliography}
}

\end{document}